\documentclass{article}

\usepackage{microtype}
\usepackage{graphicx}
\usepackage{subfigure}
\usepackage{booktabs} % for professional tables
\usepackage{float}
\usepackage{amssymb}
\usepackage{amsmath}

\DeclareMathOperator{\softmax}{softmax}

\usepackage{hyperref}

\usepackage[accepted]{icml2020}

\icmltitlerunning{Synthesizer: Rethinking Self-Attention for Transformer Models}

\begin{document}

\twocolumn[
\icmltitle{Synthesizer: Rethinking Self-Attention for Transformer Models}




% AUTHOR LIST AND ORDER NOT FINAL! 
\begin{icmlauthorlist}
\icmlauthor{Yi Tay}{to}
\icmlauthor{Dara Bahri}{to}
\icmlauthor{Donald Metzler}{to}
\icmlauthor{Da-Cheng Juan}{to}
\icmlauthor{Zhe Zhao}{to}
\icmlauthor{Che Zheng}{to}
\end{icmlauthorlist}

\icmlaffiliation{to}{Google Research, Mountain View, California}

\icmlcorrespondingauthor{Yi Tay}{yitay@google.com}

\icmlkeywords{Machine Learning, ICML}

\vskip 0.3in
]



\printAffiliationsAndNotice{}  % leave blank if no need to mention equal contribution

\begin{abstract}
The dot product self-attention is known to be central and indispensable to state-of-the-art Transformer models. But is it really required? This paper investigates the true importance and contribution of the dot product-based self-attention mechanism on the performance of Transformer models. Via extensive experiments, we find that (1) random alignment matrices surprisingly perform quite competitively and (2) learning attention weights from token-token (query-key) interactions is useful but not that important after all. To this end, we propose \textsc{Synthesizer}, a model that learns synthetic attention weights without token-token interactions. In our experiments, we first show that simple Synthesizers achieve highly competitive performance when compared against vanilla Transformer models across a range of tasks, including machine translation, language modeling, text generation and GLUE/SuperGLUE benchmarks. When composed with dot product attention, we find that Synthesizers consistently outperform Transformers. Moreover, we conduct additional comparisons of Synthesizers against Dynamic Convolutions, showing that simple Random Synthesizer is not only $60\%$ faster but also improves perplexity by a relative $3.5\%$. Finally, we show that simple factorized Synthesizers can outperform Linformers on encoding only tasks. 
\end{abstract}

\section{introduction}

Transformer models \citep{vaswani2017attention} have demonstrated success across a wide range of tasks. This has resulted in Transformers largely displacing once popular auto-regressive and recurrent models in recent years. At the heart of Transformer models lies the query-key-value dot product attention.  The success of Transformer models is widely attributed to this self-attention mechanism since fully connected token graphs, which are able to model long-range dependencies, provide a robust inductive bias.

But is the dot product self-attention really so important? Do we need it? Is it necessary to learn attention weights  via pairwise dot products? This paper seeks to develop a deeper understanding of the role that the dot product self-attention mechanism plays in Transformer models. 

The fundamental role of dot product self-attention is to learn self-alignment, i.e., to determine the relative importance of a single token with respect to all other tokens in the sequence. To this end, there have been memory metaphors and analogies constructed to support this claim. Indeed, the terms \textit{query}, \textit{keys}, and \textit{values} imply that self-attention emulates a content-based retrieval process which leverages pairwise interactions at its very core. 

Moving against convention, this paper postulates that we cannot only do without dot product self-attention but also content-based \textit{memory-like} self-attention altogether. Traditionally, attention weights are learned at the instance or sample level, where weights are produced by instance-level pairwise interactions. As a result, these instance-specific interactions often fluctuate freely across different instances as they lack a consistent global context.
 
This paper proposes \textsc{Synthesizer}, a new model that learns to synthesize the self-alignment matrix instead of manually computing pairwise dot products. We propose a diverse suite of synthesizing functions and extensively evaluate them. We characterize the source information that these synthesizing functions receive, i.e., whether they receive information from individual tokens, token-token interactions, and/or global task information. Intuitively, different source inputs to the synthesizing functions should capture diverse views, which may be useful when employed in conjunction.
 
Aside from generalizing the standard Transformer model, we show that it is possible to achieve competitive results with fully global attention weights that do not consider token-token interactions or any instance-level (local) information at all. More specifically, a \emph{random} matrix \textsc{Synthesizer} model achieves a $27.27$ BLEU score on WMT 2014 English-German\footnote{The originally reported result is $27.30$.}. Via a set of rigorous experiments, we observe that the popular and well-established dot-product content-based attention can be approximated with simpler variants such as random matrices or dense layers without sacrificing much performance in some cases. 

In our experiments, we also show that our relatively simple Synthesizer models also outperform Dynamic Convolutions \citep{wu2019pay} with a +3.5$\%$ relative improvement in perplexity while being $60\%$ faster. On encoding tasks, our factorized Synthesizers can outperform other low-rank efficient Transformer models such as Linformers \citep{wang2020linformer}.

While simple Synthesizer models are able to perform competitively, our experiments show that the pairwise dot product is still ultimately helpful. When composing our synthesizing functions with dot products, we find that they consistently improve the performance of Transformers. In general, we believe our findings will spur further investigation and discussion about the true role and utility of the self-attention mechanism in Transformer models.

% \textsc{Synthesizer} is completely transformation-based, only relies on simple feed-forward layers, and completely dispenses with dot products and explicit token-token interactions. To reiterate, this work moves away from the implied notion of a query-key-value memory store and shows that randomized alignment matrices are sufficient for many tasks in practice.
\paragraph{Our Contributions}
Our key contributions are described as follows:
\begin{itemize}
\item We propose Synthetic Attention, a new way of learning to attend without explicitly attending (i.e., without dot product attention or content-based attention). Instead, we generate the alignment matrix independent of token-token dependencies and explore a potpourri of parameterized functions for synthesizing attention matrices. 
\item We propose \textsc{Synthesizer}, a new model that leverages Synthetic Attention. The model performs competitive to state-of-the-art Transformer models on a wide range of language tasks, including machine translation and language modeling. 
\item Moreover, we show that (1) random learnable alignment matrices perform competitively and (2) token-token dependencies are not necessary to achieve good performance with Transformer models on certain tasks. 
\item On large-scale masked language modeling on the C4 dataset \citep{raffel2019exploring} and finetuning on SuperGLUE and GLUE benchmarks, we show that simple random Synthesizers can outperform/match Lightweight Dynamic convolutions \citep{wu2019pay} along with outperforming Transformers and Universal Transformers \citep{dehghani2018universal}. On two encoding tasks, factorized random Synthesizers outperform low-rank Linformers \citep{wang2020linformer}.
\end{itemize}

\section{Related Work}
Attention-based models are used across a wide spectrum of problem domains. Such models are especially popular, due to their effectiveness, in the language and vision domains. Attention models can be traced back to the machine translation models of \citep{bahdanau2014neural} and \citep{luong2015effective}, where attention is employed to learn soft word alignments between language pairs. The intuition behind the attention mechanism is deeply-rooted in the notion of memory-based retrieval \citep{graves2014neural,weston2014memory}, in which soft differentiable addressing of memory was initially proposed. 

% Learning soft alignments, also known as cross attention, have been extremely vital to problem domains such as machine translation \citep{luong2015effective}, natural language inference \citep{parikh2016decomposable} and question answering \citep{wang2016machine}. 

The paradigm of learning self-alignments, also known as self-attention, has been largely popularized by Transformer models \citep{vaswani2017attention}. This technical narrative has also been explored by a number of other recent studies, including those on intra-attention \citep{parikh2016decomposable}, self-matching networks \citep{wang2017gated}, and LSTMN \citep{cheng2016long}. To this end, Transformer models, which function primarily based on self-attention and feed-forward layers, generally serve as a reliable replacement for autoregressive recurrent models. 

The self-attention layer itself has been the subject of many recent technical innovations. For example, recent studies have investigated improving the layer's overall efficiency via sparsification and reducing the complexity of computing the alignment matrix \citep{child2019generating,kitaev2020reformer,huang2018music,tay2020sparse,Beltagy2020Longformer}. These methods are tightly coupled with the query-key-value paradigm, employing a form of memory-based content retrieval as an attention mechanism. On the other end of the spectrum, there have been studies that advocate for replacing self-attention with convolution \citep{wu2019pay}. The recent surge in interest in simplifying the attention mechanism raises important questions about the role and  utility of the pairwise dot products, which are one the defining characteristics of self-attention models. Meanwhile, in the image domain, \citep{cordonnier2019relationship} shows connection of Transformers with CNNs.

Our work is a new take on the self-attention mechanism in Transformer models. We delve deeper, starting with replacing the pairwise dot products with what we call synthesizing functions that learn attention matrices that may or may not depend on the input tokens. The most closely related work is (\citep{raganato2020fixed}), in which the authors propose using fixed (i.e., not learned) attention patterns in Transformer encoders. However, the scope of their work is limited to encoders and relies on manually defined handcrafted patterns that seem to work well. Our work takes this intuition further and expands on this narrative.

\paragraph{MLP-Mixers are Random Synthesizers}
This is an update\footnote{This paper's draft first went out a year ago, on May 2020.} discussing the relationship between Random Synthesizers and recent MLP-Mixers \citep{tolstikhin2021mlp}. There have been recent work (April 2021) that proposed All-MLP architectures for vision. Although, this work made it's appearance first in May 2020, a year before the MLP-Mixer was proposed, we show that Random Synthesizers are a form of MLP-Mixers. Random Synthesizers apply a weight matrix $R$ on the length dimension. $R$ is a $L \times L$ matrix and can be seen as a form of projection across the length dimension. This is equivalent to transposing the axis before linear projection in the token-mixer in the MLP-Mixer model. The key difference here is that (1) we use a softmax normalization on the kernel (weights) and (2) Random Synthesizers are a form of multi-headed MLP-Mixers.

\section{The Proposed Method}
\begin{figure*}[t]
\centering
     \includegraphics[width=0.8\linewidth]{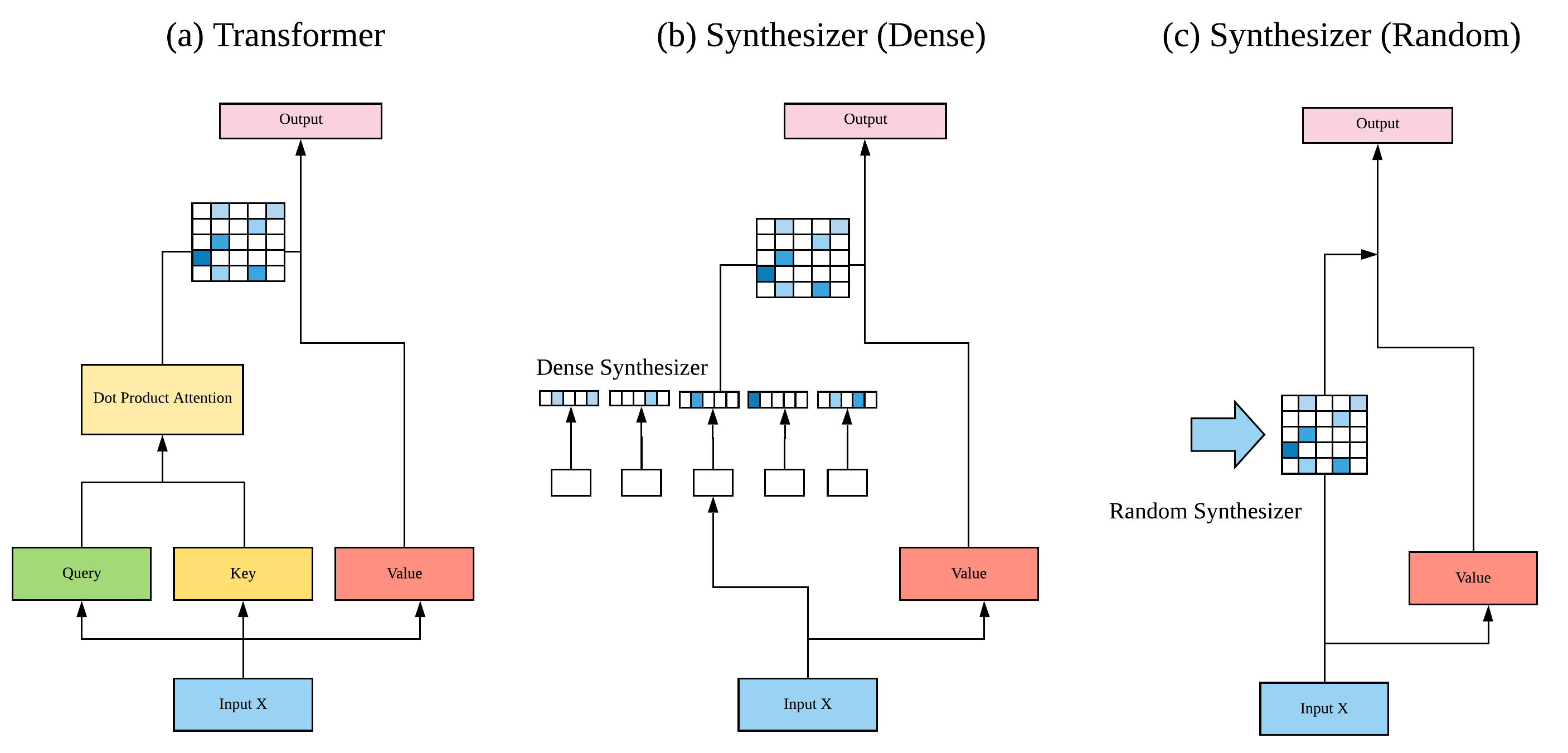}
    \label{fig:architecture}
    \caption{Our proposed \textsc{Synthesizer} model architecture.}
\end{figure*}

This section introduces our proposed \textsc{Synthesizer} model. At its core, our model is essentially a Transformer model with self-attention modules replaced with our Synthetic Attention modules. Figure \ref{fig:architecture} illustrates the key ideas behind (a) Transformer (b) Dense Synthesizers and (c) Random Synthesizers.

\subsection{Synthesizer Model}
This section introduces Synthetic Attention, our proposed self-attention module. Our model removes the notion of query-key-values in the self-attention module and directly synthesizes the alignment matrix instead. For simplicity, we describe the per head and per layer computation, which is denoted by $h$ and $\ell$ respectively in most cases.

\paragraph{Dense Synthesizer} Let us consider the simplest variation of the \textsc{Synthesizer} model which is conditioned on each input token. Overall, our method accepts an input $X_{h,\ell} \in \mathbb{R}^{N \times d}$ and produces an output of $Y_{h,\ell} \in \mathbb{R}^{N \times d}$. Here, $\ell$ refers to the sequence length and $d$ refers to the dimensionality of the model. We first adopt $F_{h,\ell}(.)$, a parameterized function, for projecting input $X_i$ from $d$ dimensions to $N$ dimensions.
\begin{align}
B_{i,h,\ell} = F_{h,\ell}(X_{i,h,\ell})     
\end{align}
where $F_{h,\ell}(.)$ is a parameterized function that maps $\mathbb{R}^{d}$ to $\mathbb{R}^{\ell}$ and $i$ is the $i$-th token of $X_{h,\ell}$ and is applied position-wise (to each vector in the sequence of length $N$). Intuitively, this can be interpreted as learning a token-wise projection to the sequence length $N$. Essentially, with this model, each token predicts weights for each token in the input sequence. In practice, we adopt a simple two layered feed-forward layer with ReLU activations for $F_{h,\ell}(.)$:
\begin{align}
F_{h,\ell}(X_{i,h,\ell}) = W_{2,h,\ell}(\sigma_{R}(W_{1,h,\ell}(X_{i,h,\ell}))   
\end{align}
where $\sigma_R$ is the ReLU activation function and $W_{1,h,\ell} \in \mathbb{R}^{d \times d}$ and $W_{2,h,\ell} \in \mathbb{R}^{d \times \ell}$. Hence, $B_{i,h,\ell}$ is now of $\mathbb{R}^{\ell}$. Given $B_{i,h,\ell} \in \mathbb{R}^{N \times N}$, we now compute:
\begin{align}
Y_{h,\ell} = \softmax(B_{h,\ell})G_{h,\ell}(X_{h,\ell})    
\end{align}
where $G_{h,\ell}(.)$ is another parameterized function of $X$ that is analogous to $V_{h,\ell}$ (value) in the standard Transformer model. This approach eliminates the dot product attention $Y = \softmax(Q_{h,\ell}K_{h,\ell}^\top)V_{h,\ell}$ altogether by replacing  $Q_{h,\ell}K_{h,\ell}^\top$ in standard Transformers with the synthesizing function $F_{h,\ell}(.)$. 

\paragraph{Random Synthesizer}
The previous variant learns synthetic attention by conditioning on each input of $X$ and projecting to $N$ dimensions. Hence, the Dense Synthesizer conditions on each token independently, as opposed to pairwise token interactions in the vanilla Transformer model. We consider another variation of \textsc{Synthesizer} where the attention weights are not conditioned on any input tokens. Instead, the attention weights are initialized to random values. These values can then either be trainable or kept fixed (denoted as \textit{Fixed}).

Let $R_{h,\ell}$ be a randomly initialized matrix. The Random Synthesizer is defined as:
\begin{align}
Y_{h,\ell} = \softmax(R_{h,\ell})G_{h,\ell}(X_{h,\ell}).    
\end{align}
where $R_{h,\ell} \in \mathbb{R}^{N \times N}$. Notably, each head adds $N^2$ parameters to the network. The basic idea\footnote{We were not expecting this variation to work at all, but it turns out to be a strong baseline.} of the Random Synthesizer is to not rely on pairwise token interactions or any information from individual token but rather to learn a task-specific alignment that works well globally across many samples. This is a direct generalization of the recently proposed fixed self-attention patterns \cite{raganato2020fixed}.

\paragraph{Factorized Models} 
The Dense Synthesizer adds $d \times N$ parameters to the network. On the other hand, the Random Synthesizer adds $N \times N$ parameters. Here, note that we omit the $Q,K$ projections in the standard Transformer which results in further parameter savings. Despite these savings, synthesized models can be cumbersome to learn when $\ell$ is large. Hence, we propose factorized variations of the \textsc{Synthesizer} models and show that these variants perform comparably in practice.
\paragraph{Factorized Dense Synthesizer}
Factorized outputs not only slightly reduce the parameter cost of the \textsc{Synthesizer} but also aid in preventing overfitting. The factorized variant of the dense synthesizer can be expressed as follows:
\begin{align}
A_{h,\ell}, B_{h,\ell} = F_{A,h,\ell}(X_{i,h,\ell}), F_{B,h,\ell}(X_{i,h,\ell})    
\end{align}
where $F_{A,h,\ell}(.)$ projects input $X_{i,h,\ell}$ into $a$ dimensions, $F_{B,h,\ell}(.)$ projects $X_{i,h,\ell}$ to $b$ dimensions, and $a \times b=N$. The output of the factorized module is now written as:
\begin{align}
Y_{h,\ell} = \softmax(C_{h,\ell})G_{h,\ell}(X_{h,\ell}).    
\end{align}
where $C_{h,\ell}=H_{A}(A_{h,\ell})* H_B(B_{h,\ell})$ where $H_A, H_B$ are tiling functions and $C_{h,\ell} \in \mathbb{R}^{N \times N}$. The tiling function simply duplicates the vector $k$ times, i.e., $\mathbb{R}^{N} \rightarrow \mathbb{R}^{N \times k}$. In this case, $H_A(\cdot)$ is a projection of $\mathbb{R}^{a} \rightarrow \mathbb{R}^{a \times b}$ and $H_B(\cdot)$ is a projection of $\mathbb{R}^{b} \rightarrow \mathbb{R}^{b \times a}$. To avoid having similar values within the same block, we compose the outputs of $H_{A}$ and $H_{B}$.

\paragraph{Factorized Random Synthesizer}
Similar to Factorized Synthesizers, we are also able to factorize $R_{h,\ell}$ into low rank matrices $R_{1,h,\ell},R_{2,h,\ell} \in \mathbb{R}^{N \times k}$.
\begin{align}
Y_{h,\ell} = \softmax(R_{1,h,\ell}R_{2,h,\ell}^{\top})G_{h,\ell}(X_{h,\ell}).    
\end{align}
Therefore, it is easy to see that, for each head, this reduces the parameter costs from $N^2$ to $2(N k)$ where $k << N$ and hence helps prevent overfitting. In practice, we use a small value of $k=8$. 

% \subsection{Overall Model Architecture}
% The remainder of the model architecture remains similar to the Transformer model. We use multi-headed Synthesizer models, two layered feed-forward layers (with layer norm and residuals). The multi-headed synthesizer module can be written as:
% \begin{align*}
% F(X) = F_{H}([Y^1(X);Y^2(X) \cdots Y^{N_H}(X)])\:\:\: \forall i \in [0, N_H]
% \end{align*}
% where $Y^i$ is the synthesizer function for head $i$ and $N_H$ is a user specified number of attention heads.

\paragraph{Mixture of Synthesizers} Finally, we note that all of the proposed synthetic attention variants can be mixed in an additive fashion. This can be expressed as:
\begin{align*}
Y_{h,\ell} = \softmax(\alpha_{1,h,\ell} S_{1,h,\ell}(X_{h,\ell})+ \\ 
\cdots \alpha_{N,h,\ell} S_{N,h,ell}(X_{h,\ell}))G_{h,\ell}(X_{h,\ell}).   
\end{align*}
where $S(.)$ is a parameterized synthesizing function and the $\alpha$ (where $\sum \alpha = 1$) are learnable weights. In the case of mixing Random Factorized with standard Dense Synthesizers, this is expressed as:
\begin{align*}
Y_{h,\ell} = \softmax(\alpha_{1,h,\ell} R_{1,h,\ell}R_{2,h,\ell}^{\top}+ \\ \alpha_{2,h,\ell}F_{h,\ell}(X_{h,\ell}))G_{h,\ell}(X).    
\end{align*}
We investigate several Mixture of Synthesizers variants in our experiments.
\paragraph{On Parameters Depending on Sequence Length}
Random and dense Synthesizers both rely on parameters that depend on length $\ell$. In general, we define a maximum length and dynamically truncate to the actual length of each batch. We note that this is in similar spirit to trainable positional encodings which have been common practice in Transformer models. Hence, we do not forsee any issue here. In the case that this is really a problem, one potential solution is to project to a smaller value $b$ and tile $b$ to the maximum sequence length. We leave this exploration to future work.

\subsection{Discussion}
This paper asks fundamental questions about the attention matrix $A$ and whether it is possible to synthesize $A$ by alternate means other than pairwise attention. It is worth noting that the regular dot product attention can also be subsumed by our \textsc{Synthesizer} framework, i.e., \textsc{Synthesizer} generalizes the Transformer model. In the case of the Transformer, the synthesizing function in question is $S(X)=F_Q(X)F_K(X)^\top$.  
\begin{table*}

\small
    \centering
    \begin{tabular}{l|ccccc}
    \toprule
    \midrule
        Model &  $S(X)$ & Condition On & Sample & Interact & $|\theta|$\\
        \midrule
        Dot Product & $F_Q(X)F_K(X_i)^\top$ & $X_j \:\: \forall j$ & Local & Yes &  $2d^2$\\
        \midrule
        Random & $R$ & N/A & Global & No & $N^2$\\
        Fac. Random & $R_1R_2^\top$ &  N/A & Global & No & $2N k$\\
         Dense & $F_1\sigma(F_2(X_i))$ & $X_i$ & Local & No & $d^2 + dN$\\
          Fac. Dense & $H_{A}(F_A(X_i))) * H_{B}(F_B(X_i)))$ & $X_i$ & Local & No & $d^2 + d(k_1+k_2)$\\
         \bottomrule
    \end{tabular}
    \caption{Overview of all Synthesizing Functions.}
    \label{tab:discussion}
\end{table*}
Table \ref{tab:discussion} lists the different model variants explored within our \textsc{Synthesizer} framework. The 'condition on' column refers to whether the synthesized output is produced as a function of $X_i$ or every $X_i, X_j$ pair. The `sample` column indicates whether a given variant leverages local or global context. Random Synthesizers are global because they share the same global alignment patterns across all samples. Dense Synthesizers are considered to be local as they are conditioned on $X_i$, which makes the alignment pattern dependent on each individual sample. To this end, it is imperative for synthesized models to have multiple heads to be effective.

\section{Experiments}
This section outlines our experimental setup and results. We first conduct experiments on five tasks to evaluate the effectiveness\footnote{Note that we are primarily interested in making controlled comparisons instead of going for the state-of-the-art result on each task.} of different Synthesizer variants along with how they compare to the vanilla Transformer. Specifically, we conduct experiments on (1) machine translation (EnDe, EnFr) (2) autoregressive language modeling (LM1B) (3) text generation (summarization and dialogue modeling and (4) multi-task natural language processing (GLUE/SuperGLUE). Details of each experiments can be found in the appendix. 

\paragraph{Notation of Variants} We use R to denote Random, D to denote Dense and V to denote vanilla dot product attention. Fix to represent Fixed Random, FR to represent Factorized Random and FD to represent Factorized random. For Mixture Synthesizers, we use + to denote that two methods are mixed.

\subsection{Comparing Synthesizer Variants and Transformer Models}
This section dives into a detailed study of multiple Synthesizer variants and the base Transformer model. 

\begin{table*}[t]
\centering
\small
    \begin{tabular}{l|ccc|cc}
    \toprule
     &    \multicolumn{3}{c}{NMT (BLEU) } & \multicolumn{2}{c}{LM (PPL)}\\
         Model & $|\theta|$ & EnDe & EnFr & $|\theta|$ & LM \\
         \midrule
         Transformer$^\dagger$
          & 67M &  27.30 & 38.10 & - &- \\
         Transformer & 67M & 27.67  & 41.57 & 70M & 
         38.21\\
        %  Transformer (Control) & 73M &  27.97 & 41.83  & - & - \\
        %  \\
         \midrule
        %  \multicolumn{6}{c}{\textbf{Synthesizer Models}} \\ 
        %  \midrule
        %  Random   && 250K& 23.89 \\
            Synthesizer (Fixed Random)  & 61M & 23.89  & 38.31 & 53M &50.52 \\
         Synthesizer (Random) & 67M & 27.27 & 41.12 & 58M & 40.60 \\
         Synthesizer (Factorized Random) & 61M & 27.30 & 41.12 &53M & 42.40 \\
        %   Synthesizer (Factorized Random Memory) &  62M &  &  \\
        %   Random Attention (Trainable) & & 500K & 27.50  \\
         Synthesizer (Dense)  & 62M & 27.43  & 41.39 & 53M & 40.88 \\
      Synthesizer (Factorized Dense) & 61M & 27.32 & 41.57  & 53M & 41.20  \\
        %  Synthesizer (Conv) & 61M & 27.56 & 41.10\\
         Synthesizer (Random + Dense) & 67M & 27.68 &  41.21 & 58M &42.35 \\ 
        Synthesizer (Dense + Vanilla) & 74M &27.57  & 41.38 & 70M & \textbf{37.27} \\
        %  Synthesizer (Factorized Random + Vanilla) & 73M & 28.16 & 41.78 & &   \\
         Synthesizer (Random + Vanilla) & 73M & \textbf{28.47} &  \textbf{41.85} & 70M & 40.05\\
         \bottomrule
    \end{tabular}
    \caption{Experimental Results on WMT'14 English-German, WMT'14 English-French Machine Translation tasks and Language Modeling One Billion (LM1B). $\dagger$ denotes original reported results in \citep{vaswani2017attention}.}
     \label{tab:ende}
     \end{table*}

\paragraph{Experimental Results on MT/LM} First, we observe that our Random Synthesizer baseline achieves $27.27$ on EnDe and $41.12$ on EnFr.  The non-trainable (i.e., fixed) variant performs substantially worse, but still yields surprisingly strong $\approx 24$ BLEU with fixed random attention weights. Most other \textsc{Synthesizer} variants achieve competitive performance, although with slight performance degradation compared to Transformers. An interesting finding is that the Mixture model of Random + Dense synthesizer performs comparably to vanilla Transformers on EnDe. When mixing the standard dot product attention, performance further increases by $+0.8$ BLEU points (EnDe). In general, the performance of \textsc{Synthesizer} variants are competitive with Transformers for this task. On LM1b, We find that the Random Synthesizers perform within $1$-$2$ PPL points away from the vanilla Transformer model. The best performing model is the Synthesizer (D+V), which achieves the best performance on this setting.

\paragraph{Results on Text Generation} 
For summarization, we find that the (R) and (D) variants do not outperform Transformers. The performance of the (D) model is $\approx 2$ Rouge-L points below Transformers. Hence, we postulate that the local sample-wise pairwise interactions are important for the summarization task. On the other hand, the utility of synthesized attention can also be observed, i.e., the (R+V) and (R+D) models both outperform Transformers. On the dialogue task, Synthesizers (R) and (D) both outperform vanilla Transformers by a reasonable margin ($\approx$ 1-3) points across most/all metrics. The best performing model here is the (D) variant. Surprisingly, unlike most other tasks, the (+V) variants do not perform well, signifying that dot product self-attention may actually be harmful for this task.

\begin{table}[H]
\small
    \centering
    \begin{tabular}{l|c|cccc}
    \toprule
    & \multicolumn{1}{c}{Sum.} & \multicolumn{4}{c}{Dialogue} \\
        Model &	 RL & B$_4$ & RL & Met. & CIDr \\
        \midrule
Trans. &35.77 & 3.20 & 13.38 & 5.89 & 18.94  \\
\midrule
\multicolumn{6}{c}{\textbf{Synthesizer Models}} \\ 
\midrule
R & 33.10 & 2.25 & 15.00 & 6.42 & 19.57  \\
D	&  33.70  & \textbf{4.02} & \textbf{15.22} & \textbf{6.61} & \textbf{20.54}  \\
D+V &  \textbf{36.02} & 3.57 &  14.22 & 6.32 & 18.87  \\ 
R+V &	 	35.95 & 2.28 & 14.79 & 6.39& 19.09  \\
\bottomrule
    \end{tabular}
    \caption{Experimental results on Abstractive Summarization (CNN/Dailymail) and Dialogue Generation (PersonaChat). We report on RL (Rouge-L), B4 (Bleu-4), Met. (Meteor) and CIDr. }
    \label{tab:gen}
\end{table}

\paragraph{Comparing Synthesizers with Dynamic Convolutions}
To ascertain the competitiveness of Synthesizers, we also compare them with Dynamic convolutions \citep{wu2019pay}. We compare them on (1) pretraining perplexity using the masked language modeling objective on C4 and (2) downtream finetuning results on GLUE and SuperGLUE.

\paragraph{Results on Masked Language Modeling}
We also benchmark the speed of these models. In order to do so, we conduct additional experiments on the T5 adaptation of masked language modeling on the C4 dataset \citep{raffel2019exploring} by comparing against lightweight dynamic convolutions \citep{wu2019pay} on a masked language modeling task. We also take this chance to benchmark the speed of Synthesizers compared with Transformers. Experiments are conducted on Mesh Tensorflow \citep{shazeer2018mesh} and ran on 2x2 TPU V3 Chips for approximately $524K$ steps.
\begin{table}[]
    \centering
    \small
    \begin{tabular}{l|cccc}
    \toprule
       Model  &  Log PPL & Steps/Sec & Params & TFLOPS\\
       \midrule
        Trans. & 1.865 & 3.90 & 223M & $3.70$\\ 
        % Universal Transformer \citep{dehghani2018universal}  & 2.111 & 0.93 & 84M & \\
        DyConv  & 2.040 & 2.65 & 257M & $3.93$\\ 
        LightConv & 1.972 & 4.05 & 224M & $3.50$ \\ \hline
        Syn (D) & 1.965 & 3.61 & 224M  & $3.80$\\ 
        Syn (R) & 1.972 & \textbf{4.26} & 254M & $3.36$ \\ 
        Syn (R+V) & 1.849 & 3.79 & 292M & $4.03$\\
        Syn (D+V) & \textbf{1.832} & 3.34 & 243M & $4.20$\\ 
        \bottomrule
    \end{tabular}
    \caption{Validation perplexity scores on C4 dataset \citep{raffel2019exploring}. All models are at approximately similar parameterization.}
    \label{tab:ppl}
\end{table}

\paragraph{Results on MLM} Table \ref{tab:ppl} reports the validation set log perplexity on masked language modeling\footnote{Note that this follows the sequence transduction style in T5.}. We observe that Synthesizers (R) can outperform Dynamic Convolutions by a relative +3.5\% while being $+60\%$ faster. Against Lightweight Dynamic Convolutions, we match the performance while being $+5\%$ faster. Given that this is the simple random Synthesizer baseline, we find this extremely interesting how it is able to outperform dynamic convolutions, a relatively complex model. The Random Synthesizer also has less FLOPS compared to both convolution models. On the other hand, the Mixture Synthesizer models that use the dot product attention improves the performance of the base Transformer model with relatively an equal model speed. Finally, similar to the earlier results, we see a consistent performance gain of Synthesizer (D+V) and Synthesizer (R+V) outperforming the base Transformer model.
\begin{table*}[t]
\small
    \centering
    \begin{tabular}{l|c|cccccccccccccc}
    \toprule
        Model &	Glue &	CoLA&	SST&	MRPC	&STSB &QQP&	MNLI&	QNLI&	RTE \\ 
        \midrule
T5 (Base)
& 83.5 &	53.1 &	\textbf{92.2} &	\textbf{92.0/88.7} &	89.1/88.9 &	88.2/91.2 &	84.7/\textbf{85.0} &	91.7 &	76.9 \\ 
T5 (Base+) & 82.8 &	54.3&	92.9&	88.0/83.8 &	85.2/85.4 &	88.3/91.2 &	84.2/84.3 &	91.4	&79.1\\
DyConv & 69.4 &	33.9 & 	90.6 & 82.6/72.5 & 60.7/63.1 & 84.2/88.2 & 73.8/75.1	& 84.4 &	58.1 \\
\midrule
Syn (R) & 75.1 & 	41.2 &	91.2 & 85.9/79.4 & 74.0/74.3 & 85.5/89.0 & 77.6/78.1 &87.6 &	59.2 \\
Syn (D) & 72.0 &	18.9 &	89.9 & 86.4/79.4 &	75.3/75.5 &	85.2/88.3 &	77.4/78.1 &	86.9& 	57.4 \\
Syn (D+V) & 82.6 &	48.6 &	92.4 &	91.2/87.7 &	88.9/89.0 &	88.6/91.5 &	84.3/84.8	&91.7 &	75.1 \\
Syn (R+V)	& \textbf{84.1} &	\textbf{53.3} &	\textbf{92.2} & 91.2/87.7 & \textbf{89.3/88.9} & \textbf{88.6/91.4} & \textbf{85.0}/84.6	& \textbf{92.3} & 	\textbf{81.2} \\
\bottomrule
    \end{tabular}
    \caption{Experimental results (dev scores) on multi-task language understanding (GLUE benchmark) for \textit{small} model and \texttt{en-mix} mixture. Note: This task has been co-trained with SuperGLUE.}
    \label{tab:glue}
    \centering
    \begin{tabular}{l|c|cccccccccccccc}
    \toprule
        Model &	SGlue &	BoolQ &	CB & 	CoPA & MultiRC &	ReCoRD &	RTE & 	WiC &	WSC \\ 
        \midrule
T5 (Base) &	70.3 &	78.2 &	72.1/83.9 &	59.0 &	73.1/32.1 &
\textbf{71.1/70.3}	& 77.3	& \textbf{65.8} &\textbf{80.8}\\
T5 (Base+) & 70.7 &	79.3 &	81.1/87.5 &	60.0 &	75.1/34.4	& 71.7/70.7 &	80.5 & 64.6	 &71.2\\ 
DyConv & 57.8 &	66.7 &	65.9/73.2&	58.0	& 57.9/8.71 &	58.4/57.4 & 	69.0 &	58.6 &	73.1 \\
\midrule
Syn (R) &	61.1 &	69.5 &	54.6/73.2 &	60.0	& 63.0/15.7 &	58.4/57.4 &	67.5 & 	64.4 &	66.3 \\
Syn (D) & 58.5 &	69.5 & 51.7/71.4 &	51.0 &	66.0/15.8	& 54.1/53.0 &	67.5 &	65.2 &	58.7 \\
Syn (D+V) & 69.7 &	79.3 &74.3/85.7	& 64.0 & 73.8/33.7 &	69.9/69.2 &	78.7 &	64.3	 &68.3 \\
Syn (R+V) &	\textbf{72.2} &	\textbf{79.3} &	\textbf{82.7/91.1} &	\textbf{64.0} &	\textbf{74.3/34.9} &	70.8/69.9	& \textbf{82.7} &	64.6 &	75.0 \\
        \bottomrule
    \end{tabular}
    \caption{Experimental results (dev scores) on multi-task language understanding (SuperGLUE benchmark) for \textit{small} model and \texttt{en-mix} mixture. Note: This task has been co-trained with GLUE.}
    \label{tab:superglue}
\end{table*}

\paragraph{Results on GLUE and SuperGLUE}
Tables \ref{tab:glue} and \ref{tab:superglue} report results on the GLUE and SuperGLUE benchmarks. We note that the (R) and (D) variants of \textsc{Synthesizer} do not achieve reasonable performance. This can be largely attributed to the fact that the encoder self-attention in the T5 setting also functions as a cross-sentence attention. For example, in the entailment or reading comprehension tasks, the premise and hypothesis are concatenated together and self-attention effectively acts as cross-sentence attention\footnote{On a related note, the perceived success of pairwise self-attention might also be attributed to the fact that these public benchmarks are bias towards pairwise matching tasks. In reality, this is computationally prohibitive for many practical real-world applications \citep{seo2018phrase}. }. On datasets like SST, a straightforward sentiment classification task, this cross sentence attention is not necessary and therefore Syn (R) and Syn (D) both perform competitively. To this end, Dynamic Convolutions \citep{wu2019pay} also do not have this encoder "cross-attention" and therefore also suffer on many of these pairwise matching tasks. Notably, in this `no cross attention' setting, the Random Synthesizers are are 4 to 5 percentage points higher in GLUE/SuperGLUE score compared to Dynamic Convolutions.

Optimistically, we observe that the mixture model Syn (R+V) outperforms the T5 model by a substantial margin (+1.9 points on SuperGLUE and +0.6 points on GLUE). Naturally, the hybrid mixture model also very substantially outperforms Dynamic Convolution. Finally to ensure that the Syn (+V) variations are not outperforming Transformers due to simply having more parameters, we also compared with T5 (Base+) which has equal number of parameters to Syn (+V) variants (approximately $\approx 10M$ more parameters). Our results show that Synthesizers (+V) still outperform T5 (Base+).

\subsection{Comparing Synthesizers with Linformers}
We conduct more experiments comparing factorized random Synthesizers with Linformers. Since Linformer cannot be used to decode, we compare them on two encoding tasks from tensorflow datasets (AGnews \citep{zhang2015character} and movie reviews \citep{maas-EtAl:2011:ACL-HLT2011}). We use $k$=$32$ for both factorized models. We also benchmark Transformers on this task. Note we do not use contextualized embeddings so results are not comparable with other work. 
\begin{table}[H]
    \centering
    \begin{tabular}{l|ccc}
    \toprule
      Model   &  News & Reviews & Steps/Sec   \\
      \midrule
     Transformer    & 88.83 & 81.34 & 1.09 \\
     Linformer  & 86.50&  82.86 & 1.09\\ 
     \midrule
    % Synthesizer (D) & 87.71 & \\
    % Synthesizer (R) & & \\
    Syn (FR) & 86.53 & 83.39 & \textbf{1.10}\\
     Syn (FR+V) & \textbf{89.13} & \textbf{84.61} & 0.80\\ 
         \bottomrule
    \end{tabular}
    \caption{Results on Encoding only tasks (accuracy).}
    \label{tab:my_label}
\end{table}
\paragraph{Results} We notice that factorized Synthesizers (FR) are competitive with Linformers and Transformers on this task. The accuracy of Syn (FR) is competitive with Linformers while Syn (FR+V) outperforms both Transformers and Linformers.

\section{Qualitative Analysis}

\begin{figure}
 \centering
     \includegraphics[width=0.5\linewidth]{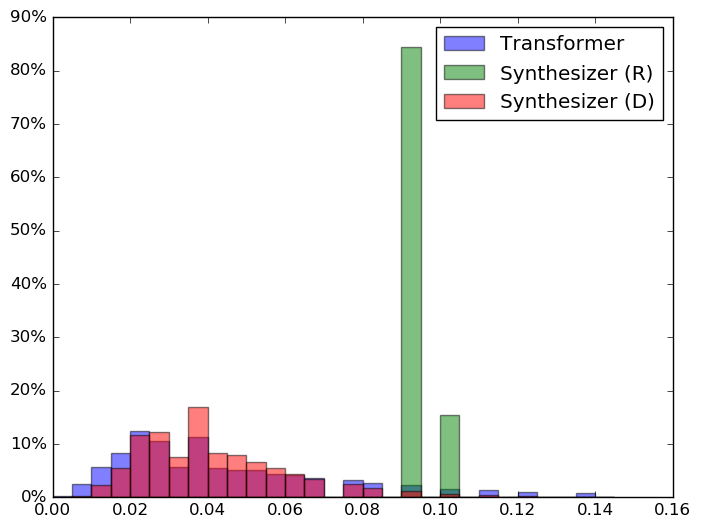}
    \caption{Init Decoder weights (Reference)}
    \label{fig:init}
\end{figure}

\begin{figure}[H]
\begin{minipage}{0.32\linewidth}
  \centering
     \includegraphics[width=1.0\linewidth]{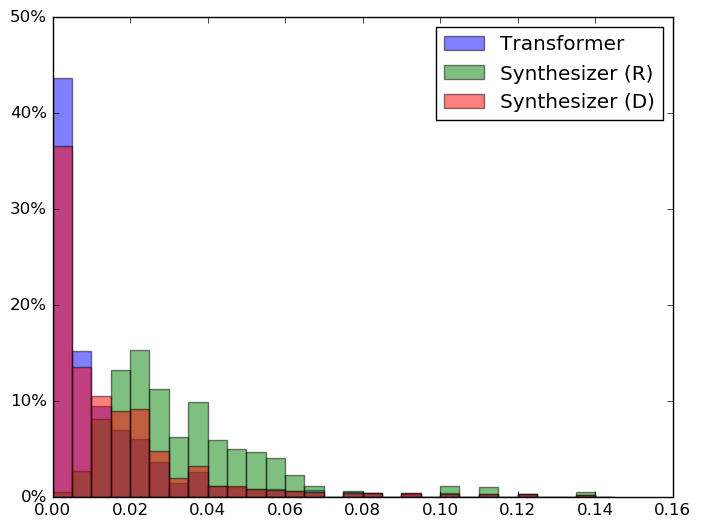}
    \\ {Enc L1}
    \label{fig:sortiter}
\end{minipage}\hfill
\begin{minipage}{0.32\linewidth}
  \centering
     \includegraphics[width=1.0\linewidth]{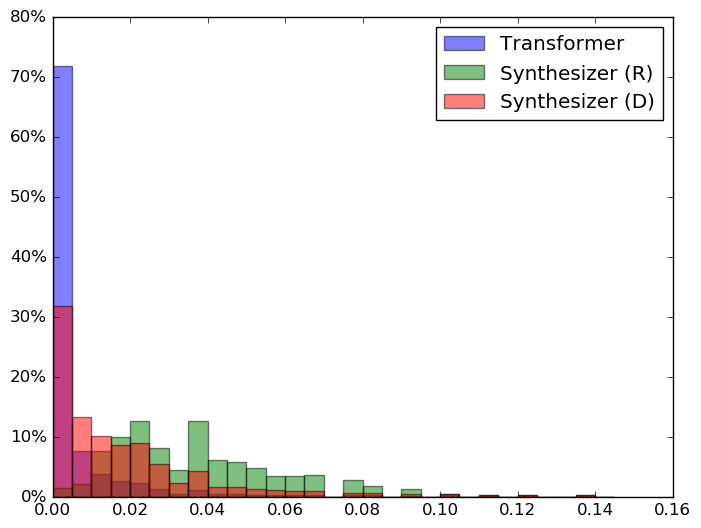}
    \\ {Enc L3}
    \label{fig:sortiter}
\end{minipage}\hfill
\begin{minipage}{0.32\linewidth}
  \centering
     \includegraphics[width=1.0\linewidth]{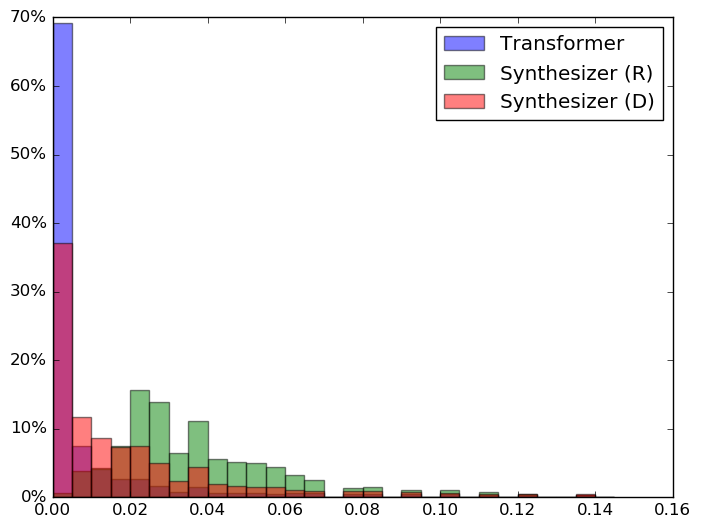}
    \\ {Enc L5}
    \label{fig:sortiter}
\end{minipage}\hfill 
\label{fig:hist1}
% \caption{Histogram of Encoder Attention Weights on MT (WMT EnDe).}
\begin{minipage}{0.32\linewidth}
  \centering
     \includegraphics[width=1.0\linewidth]{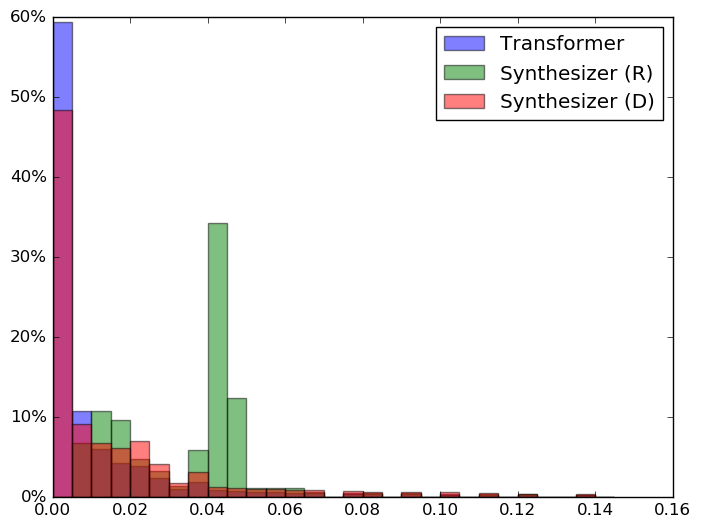}
    \\ {Dec L1}
    \label{fig:sortiter}
\end{minipage}\hfill 
\label{fig:analysis}
% \caption{Histogram of Encoder Attention Weights (pre-Softmax) on MT (WMT EnDe).}
\begin{minipage}{0.32\linewidth}
  \centering
     \includegraphics[width=1.0\linewidth]{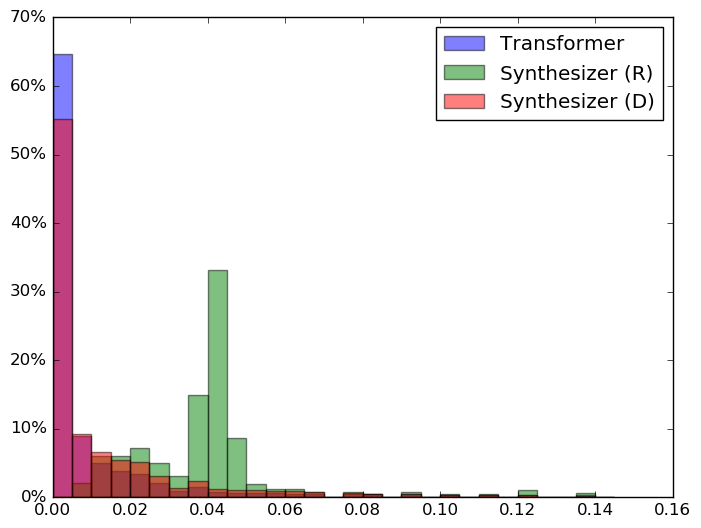}
    \\ {Dec L3}
    \label{fig:sortiter}
\end{minipage}\hfill 
\label{fig:analysis}
\begin{minipage}{0.32\linewidth}
  \centering
     \includegraphics[width=1.0\linewidth]{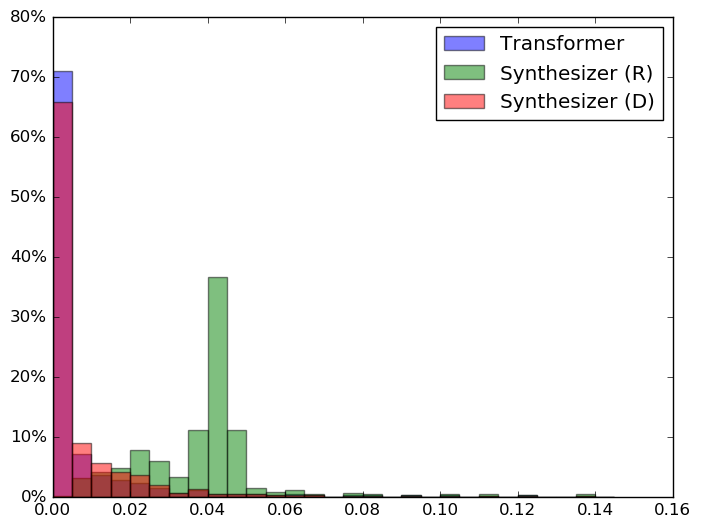}
    \\ {Dec L5}
    \label{fig:sortiter}
\end{minipage}\hfill 
\label{fig:hist2}
\caption{Histogram of Encoder and Decoder Attention Weights on MT (WMT EnDe). L denotes the layer number and Enc/Dec denotes encoder or decoder.}
\label{hist}
\end{figure}

\paragraph{Distribution of Weights} We are interested in investigating how the synthetically generated attention weights differ from the dot product attention weights. Figure \ref{hist} shows the attention histograms on trained Transformer and \textsc{Synthesizer} models. We report histograms at layers $1$, $3$, and $5$ of a 6 layered (Transformer or \textsc{Synthesizer}) model at $50K$ steps. We found that the weight distributions remain relatively identical thereafter. Figure \ref{fig:init} shows the initialization state. We observe that there are distinct differences in the weight distribution of \textsc{Synthesizer} and Transformer models. The variance of the \textsc{Synthesizer} weights tends to be higher. On the other hand, the weights on the Transformer model tends to gravitate near $0$ and have smaller variance. There are also notable differences across the (R) and (D) \textsc{Synthesizer} variants. Specifically, the (D) model in general has greater max values with more values in the $0.1$-$0.2$ range while the values of the $R$ model tends to stay closer to $0$.

\subsection{What patterns do Synthesizers learn?}
In this section, we perform a deeper analysis of the \textsc{Synthesizer} model.

\begin{figure}[H]
\begin{minipage}{0.18\linewidth}
  \centering
     \includegraphics[width=1.0\linewidth]{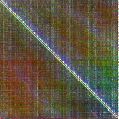}
    \\ {Vanilla}
    \label{fig:sortiter}
\end{minipage}\hfill
\begin{minipage}{0.18\linewidth}
  \centering
     \includegraphics[width=1.0\linewidth]{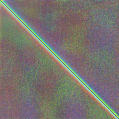}
    \\ {Random}
    \label{fig:sortiter}
\end{minipage}\hfill
\begin{minipage}{0.18\linewidth}
  \centering
    \includegraphics[width=1.0\linewidth]{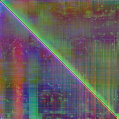}
    \\ {Fr}
    \label{fig:temperature}
\end{minipage}\hfill
\begin{minipage}{0.18\linewidth}
  \centering
     \includegraphics[width=1.0\linewidth]{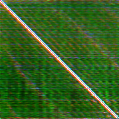}
    \\ {Dense}
    \label{fig:sortiter}
\end{minipage}
\begin{minipage}{0.18\linewidth}
  \centering
     \includegraphics[width=1.0\linewidth]{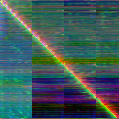}
    \\ {FD}
    \label{fig:sortiter}
\end{minipage}\hfill 
\label{fig:analysis}
\caption{Visual analysis of Synthetic Attention (encoder) on WMT EnDe.}
\end{figure}
\begin{figure}[H]
\begin{minipage}{0.18\linewidth}
  \centering
     \includegraphics[width=1.0\linewidth]{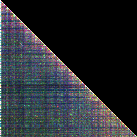}
    \\ {Vanilla}
    \label{fig:sortiter}
\end{minipage}\hfill
\begin{minipage}{0.18\linewidth}
  \centering
     \includegraphics[width=1.0\linewidth]{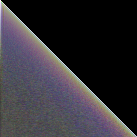}
    \\ {Random}
    \label{fig:sortiter}
\end{minipage}\hfill
\begin{minipage}{0.18\linewidth}
  \centering
    \includegraphics[width=1.0\linewidth]{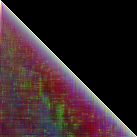}
    \\ {FR}
    \label{fig:temperature}
\end{minipage}\hfill
\begin{minipage}{0.18\linewidth}
  \centering
     \includegraphics[width=1.0\linewidth]{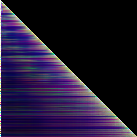}
    \\ {Dense}
    \label{fig:sortiter}
\end{minipage}
\begin{minipage}{0.18\linewidth}
  \centering
     \includegraphics[width=1.0\linewidth]{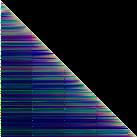}
    \\ {FD}
    \label{fig:sortiter}
\end{minipage}\hfill 
\label{fig:analysis}
\caption{Visual analysis of Synthetic Attention (decoder) on WMT EnDe.}
\end{figure}

\begin{figure}
\begin{minipage}{0.48\linewidth}
 \centering
     \includegraphics[width=0.9\linewidth]{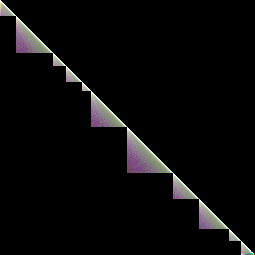}
    \caption{Synthesizer weights on LM1B.}
    \label{fig:synlm}
\end{minipage}
\begin{minipage}{0.48\linewidth}
 \centering
     \includegraphics[width=0.9\linewidth]{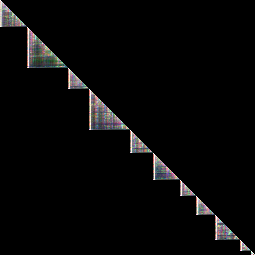}
    \caption{Transformer weights on LM1B.}
    \label{fig:synlm}
\end{minipage}
\end{figure}

\paragraph{Analysis} Finally, we are interested to understand what these Synthesizer models are learning. We inspect the random synthetic attention weights for language modeling task LM1B and visualise the differences compared to the vanilla attention. We find that, for the LM task, Synthesizers are capable of learning a local window, emulating the vanilla Transformer quite closely despite starting from completely random. The weights, however, seem smoother and less coarse as compared to the Transformer. This seems to reflect what we expect since the Synthesizer does not benefit from token specific information. We provide additional analysis and visualisation of weights for the Machine Translation task in the supplementary material.

\subsection{Overall Summary of Quantitative Results}
This section summarizes our overall findings.
\paragraph{Synthetic Attention is competitive even without Dot Product Attention}
On all evaluated tasks, we showed that synthesized attention functions competitively, i.e., it achieves performance reasonably close to the dot product self-attention. On one task (dialogue generation), the dot product self-attention is found to actually degrade performance. Amongst the other tasks, machine translation is the least affected by the removal of the vanilla dot product. These findings allow us to introspect about whether pairwise comparisons for self-attention are even necessary. On the multi-task language understanding benchmark, the self-attention functions as a form of cross-attention by concatenating sentence pairs. Hence, synthesize attention performance is considerably worse than vanilla Transformers. 
    \paragraph{Synthetic Attention and Dot Product Attention are highly complementary} 
    Overall, we also observe that the dot product attention is very helpful. To this end, synthetic attention is highly complementary to the pairwise dot product attention. While Synthetic Attention can usually achieve competitive and fast performance on its own,  synthetic attention boosts performs, composing multiple synthetic attention (and dot product attention) together shows gains on almost all tasks that we have investigated. Hence, we believe this to be a robust finding. 
    
\paragraph{The simplest Synthesizers such as Random Synthesizers are fast competitive baselines} Finally, we note that simple random Synthesizers are competitive with dynamic convolutions and Linformers, which are recently proposed models. On two encoding task and a large-scale masked language modeling task, we show that random (or factorized random) Synthesizers remain competitive to other fast or efficient Transformer models.

\section{Conclusion}
This paper proposed \textsc{Synthesizer}, a new Transformer model that employs Synthetic Attention. We conducted a principled study to better understand and evaluate the utility of global alignment and local, instance-wise alignment (e.g., independent token and token-token based) in self-attention. We show that, on multiple tasks such as machine translation, language modeling, dialogue generation, masked language modeling and document classification, synthetic attention demonstrates competitive performance compared to vanilla self-attention. Moreover, for the dialogue generation task, pairwise interactions actually hurt performance. Notably, we reemphasize that this study refers to self-attention. We found that we are not able to replace cross-attention with simpler variants in most cases. Via a set of additional large-scale experiments, also find that Synthesizers can outperform or match Dynamic Convolutions and Factorized Synthesizers can outperform other low rank Linformer models.

\nocite{langley00}

\bibliography{example_paper}
\bibliographystyle{icml2020}

% %%%%%%%%%%%%%%%%%%%%%%%%%%%%%%%%%%%%%%%%%%%%%%%%%%%%%%%%%%%%%%%%%%%%%%%%%%%%%%%
% %%%%%%%%%%%%%%%%%%%%%%%%%%%%%%%%%%%%%%%%%%%%%%%%%%%%%%%%%%%%%%%%%%%%%%%%%%%%%%%
% % DELETE THIS PART. DO NOT PLACE CONTENT AFTER THE REFERENCES!
% %%%%%%%%%%%%%%%%%%%%%%%%%%%%%%%%%%%%%%%%%%%%%%%%%%%%%%%%%%%%%%%%%%%%%%%%%%%%%%%
% %%%%%%%%%%%%%%%%%%%%%%%%%%%%%%%%%%%%%%%%%%%%%%%%%%%%%%%%%%%%%%%%%%%%%%%%%%%%%%%
% \appendix
% \section{Do \emph{not} have an appendix here}

% \textbf{\emph{Do not put content after the references.}}
% %
% Put anything that you might normally include after the references in a separate
% supplementary file.

% We recommend that you build supplementary material in a separate document.
% If you must create one PDF and cut it up, please be careful to use a tool that
% doesn't alter the margins, and that doesn't aggressively rewrite the PDF file.
% pdftk usually works fine. 

% \textbf{Please do not use Apple's preview to cut off supplementary material.} In
% previous years it has altered margins, and created headaches at the camera-ready
% stage. 
%%%%%%%%%%%%%%%%%%%%%%%%%%%%%%%%%%%%%%%%%%%%%%%%%%%%%%%%%%%%%%%%%%%%%%%%%%%%%%%
%%%%%%%%%%%%%%%%%%%%%%%%%%%%%%%%%%%%%%%%%%%%%%%%%%%%%%%%%%%%%%%%%%%%%%%%%%%%%%%

\end{document}